\pdfoutput=1

\documentclass[11pt]{article}

\usepackage[final]{acl}

\usepackage{times}
\usepackage{latexsym}
\usepackage[T1]{fontenc}

\usepackage{CJKutf8}

\usepackage[utf8]{inputenc}
\usepackage{CJKutf8}

\usepackage{microtype}
\usepackage{amsmath}

\usepackage{inconsolata}

\usepackage{graphicx}

\usepackage{xcolor}
\usepackage{colortbl}
\usepackage{booktabs}
\usepackage{adjustbox}
\usepackage{arydshln}

\usepackage{amsmath}
\usepackage{mathtools}
\usepackage{amssymb}
\usepackage{dsfont}
\usepackage{tikz}
\usepackage[edges]{forest}
\usepackage{arydshln}
\usepackage{pgfplots}
\usepackage{makecell}
\usepackage{custom}
\usepackage{CJKutf8} 
\usepackage{dsfont}
\usepackage{courier}
\usepackage[most]{tcolorbox}
\usepackage{multirow}
\newtcolorbox{taskbox}[2][]{
	enhanced, breakable,
	colframe=blue3!40,
	colback=blue5!5,
	arc=1mm,
	outer arc=1mm,
	fontupper=\small,
	fontlower=\small,
	coltitle=blue1,
	fonttitle=\bfseries,
	boxsep=1mm,
	left=0mm,
	right=0mm,
	top=0mm,
	bottom=0mm,
	before={\noindent},
	segmentation style={solid, blue3},
	title=#2,
	#1
}
\definecolor{mycustomcolor}{RGB}{210,226,210}

\definecolor{red}{RGB}{184,49,55}
\newcommand{\Red}[1]{\textcolor{red}{#1}}

\newcommand\assertion[1]{\colorbox[RGB]{237,233,195}{#1}}
\newcommand\process[1]{\colorbox[RGB]{226,236,247}{#1}}
\newcommand\result[1]{\colorbox[RGB]{236,225,230}{#1}}
\newcommand\wrong[1]{\colorbox[RGB]{227,238,225}{#1}}

\definecolor{blue}{RGB}{55,83,156}
\newcommand{\Blue}[1]{\textcolor{blue}{#1}}

\definecolor{green}{RGB}{100,141,63}
\newcommand{\Green}[1]{\textcolor{green}{#1}}
\newcommand{\modelname}{\texttt{WoT}}
\newtcolorbox{mybox}[2][]{
    width=\columnwidth,
    colback =  gray!8,
    colframe = gray!8, boxsep=0pt,left=10pt,right=10pt,top=0pt,bottom=0pt,
    fontupper=\linespread{1.1}\selectfont,
    title=#2,#1
}

\title{\Red{Wrong}-of-Thought: An Integrated Reasoning Framework with \\ Multi-Perspective Verification and Wrong Information}

\author{
	Yongheng Zhang$^{1}$~~
	Qiguang Chen$^{2}$~~
	\textbf{Jingxuan Zhou}$^{1}$~~
	\textbf{Peng Wang}$^{1}$
	\\
	\textbf{Jiasheng Si}$^{3}$~~
	\textbf{Jin Wang}$^{4}$~~
	\textbf{Wenpeng Lu}$^{3}$~~
    \textbf{Libo Qin}$^{1}$\thanks{\,\, Corresponding Author.}\\
	$^{1}$School of Computer Science and Engineering, Central South University, China\\
	$^{2}$Research Center for SCIR, Harbin Institute of Technology, Harbin, China\\
	$^{3}$Key Laboratory of Computing Power Network and Information Security, Ministry of Education\\
	Qilu University of Technology (Shandong Academy of Sciences), China\\
$^{4}$Yunnan University, Kunming, China
}

\usepackage{ragged2e}

\begin{document}
\begin{CJK}{UTF8}{gbsn}	

\maketitle

	\begin{abstract}
Chain-of-Thought (CoT) has become a vital technique for enhancing the performance of Large Language Models (LLMs), attracting increasing attention from researchers.
One stream of approaches focuses on the iterative enhancement of LLMs by continuously verifying and refining their reasoning outputs for desired quality.
Despite its impressive results, this paradigm faces two critical issues:
(1) \textit{Single verification method}: The current paradigm relies solely on a single verification method.
(2) \textit{Wrong Information Ignorance}: The traditional paradigm directly ignores wrong information during reasoning and refines the logic paths from scratch each time. 
To address these challenges, we propose Wrong-of-Thought (\modelname{}), which includes two core modules:
(1)~\textit{Multi-Perspective Verification}: A multi-perspective verification method for accurately refining the reasoning process and result, and (2)~\textit{Wrong Information Utilization}: Utilizing wrong information to alert LLMs and reduce the probability of LLMs making same mistakes. 
Experiments on 8 popular datasets and 5 LLMs demonstrate that \modelname{} surpasses all previous baselines. In addition, \modelname{} exhibits powerful capabilities in difficult computation tasks.
\end{abstract}
	\section{Introduction}
\begin{quotation}
	\textit{Failure is the mother of success.}
	\begin{flushright}
		\par \textit{-- Chinese Idiom}
	\end{flushright}
\end{quotation}

In recent years, large language models (LLMs) have made significant advancements in a series of natural language processing tasks \citep{achiam2023gpt, touvron2023llama}. Additionally, with the emergence of Chain-of-Thought (CoT) \cite{wei2022chain}, the performance of LLMs has been further unlocked by guiding them through step-by-step reasoning \citep{liu2023logicot,qin2023cross}.

\begin{figure}[t]
	\centering
	\includegraphics[width=0.49\textwidth]{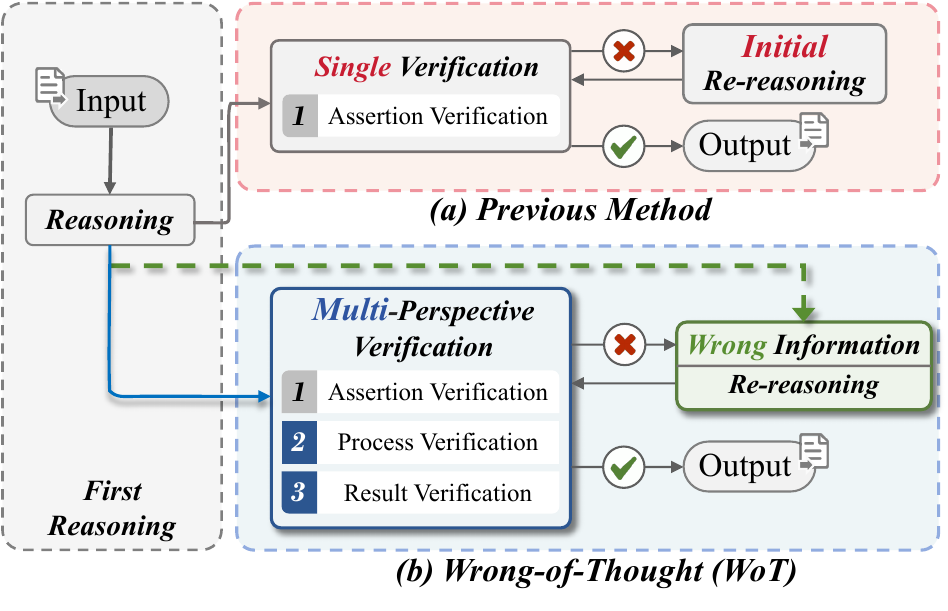}
	\caption{Previous multi-thoughts integration methods~(a) vs. Wrong-of-Thought~(b). Previous methods only used a \textit{\textbf{\Red{Single}} Verification} and did not utilize the wrong information. In contrast, \modelname{} offers \textit{\textbf{\Blue{Multi}}-Perspective Verification} and utilizes \textit{\textbf{\Green{Wrong}} Information}.}
	\vspace{-0.1cm}
	\label{intro}
\end{figure}

A common category of CoT research focuses on iteratively enhancing LLMs by continuously re-verifying and refining corresponding reasoning outputs to achieve better quality. 
\citet{madaan2023self} demonstrate this approach by prompting the model to self-verify the results and provide feedback on previously generated drafts, producing better outputs.
Similarly, \citet{chen2023teaching} improve code debugging by leveraging external program execution results and model explanation code.
When examining methodologies to guide the rethinking of models, \citet{zheng2023progressive} emphasize the reuse of previously generated answers.
Meanwhile, \citet{qi2023art} introduce a problem-solving framework inspired by human divide-and-conquer strategies, which incorporates self-questioning and recursive thinking processes.
Building upon this, \citet{liu2023plan} propose XoT, shown as Figure~\ref{intro}~(a), which integrates multiple reasoning paths with multiple logical modes. Specifically, they generate the rationale in program format and apply a single verification method to check the correctness of the reasoning. If errors are detected, the LLMs are instructed to switch to another reasoning thought and start the reasoning process from scratch.
Despite achieving impressive results, they still face two significant challenges:
\begin{itemize}
\item [(1)] \textit{Single verification method}: They rely solely on the single verification method like basic syntax assertions, resulting in errors that fail to fully evaluate and validate the reasoning of models. This approach leads to suboptimal verification accuracy, significantly impeding overall reasoning performance.
\item [(2)] \textit{Wrong Information Ignorance}: 
Once the error is detected, they typically disregard wrong information and re-generate the reasoning from scratch. However, it also loses a large amount of feedback signals brought by error information, which is often considered very important in model understanding~\citep{zhang2024context,tong2024can, chen2024boosting}. 
\end{itemize}

Motivated by this, we introduce the Wrong-of-Thought (\modelname{}) framework, as illustrated in Figure~\ref{intro}~(b). To address the first challenge, we introduce \textit{Multi-Perspective Verification}, which incorporates two additional explicit verification methods, mirroring human problem-solving processes. First, it ensures the variables in equations or code match the information provided in the question. Second, it resolves again the question to check for consistency in the results. We instruct LLMs to integrate these two perspectives to enhance solution verification.
To address the second challenge, we introduce \textit{Wrong Information Utilization}, which utilizes previous wrong reasoning information to guide LLMs in avoiding similar mistakes. By referencing past mistakes, LLMs can enhance their reasoning performance and minimize repetitive errors.

Experiments are conducted on 8 datasets and 5 LLMs. The results indicate that the \modelname{} performs exceptionally well across all benchmark tests, surpassing all existing baselines. Furthermore, in-depth analytical experiments demonstrate that \modelname{} excels at difficult computational tasks.

The key contributions of this work are:
\begin{itemize}
\item [(1)] We first point out two main drawbacks in iterative reasoning, which lie in the monotonous verification perspective and the ignorance of wrong information feedback for ultimate limited logical improvement.
\item [(2)] We introduce Wrong-of-Thought (\modelname{}) to solve these drawbacks, which mainly consists of two modules: \textit{Multi-Perspective Verification} and \textit{Wrong Information Utilization}. These modules enable accurate verification and effective utilization of wrong information.
\item [(3)] Our experiments on 8 datasets and 5 LLMs have shown that \modelname{} achieves superior performance.  In addition, \modelname{} demonstrates strong problem-solving abilities in questions involving difficult mathematical reasoning.
\end{itemize}

All code will be open-sourced and publicly available at \url{https://github.com/BRZ911/Wrong-of-Thought}.

\begin{figure}[t]
	\centering
	\includegraphics[width=0.49\textwidth]{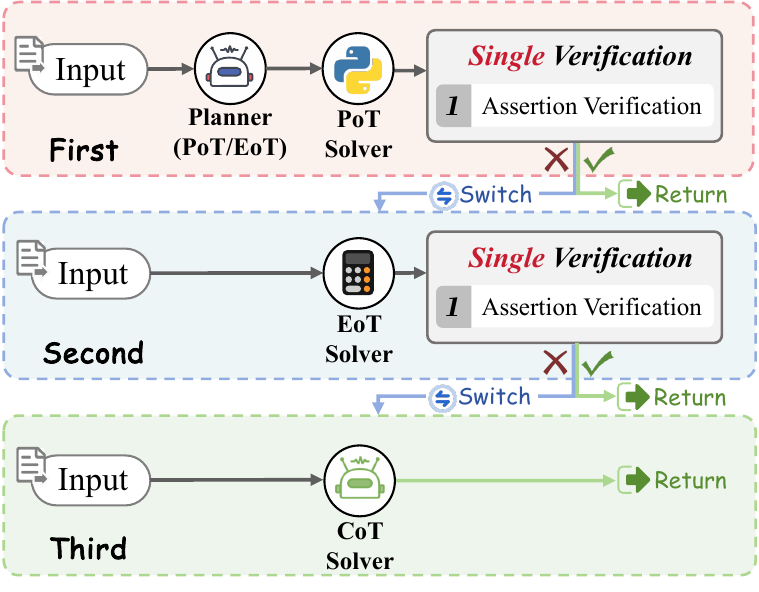}
	\caption{XoT Framework. First, select a reasoning method, either PoT or EoT and then apply assertion verification to make a judgment. If the reasoning is found to be incorrect, switch to the alternative method and restart the reasoning. Verify again, and if the verification is correct, return the answer. If the reasoning reaches the third step, utilize CoT reasoning as the answer.}
	\label{back}
\end{figure}
	
\section{Preliminary}

\begin{figure*}[t]
	\centering
	\includegraphics[width=160mm]{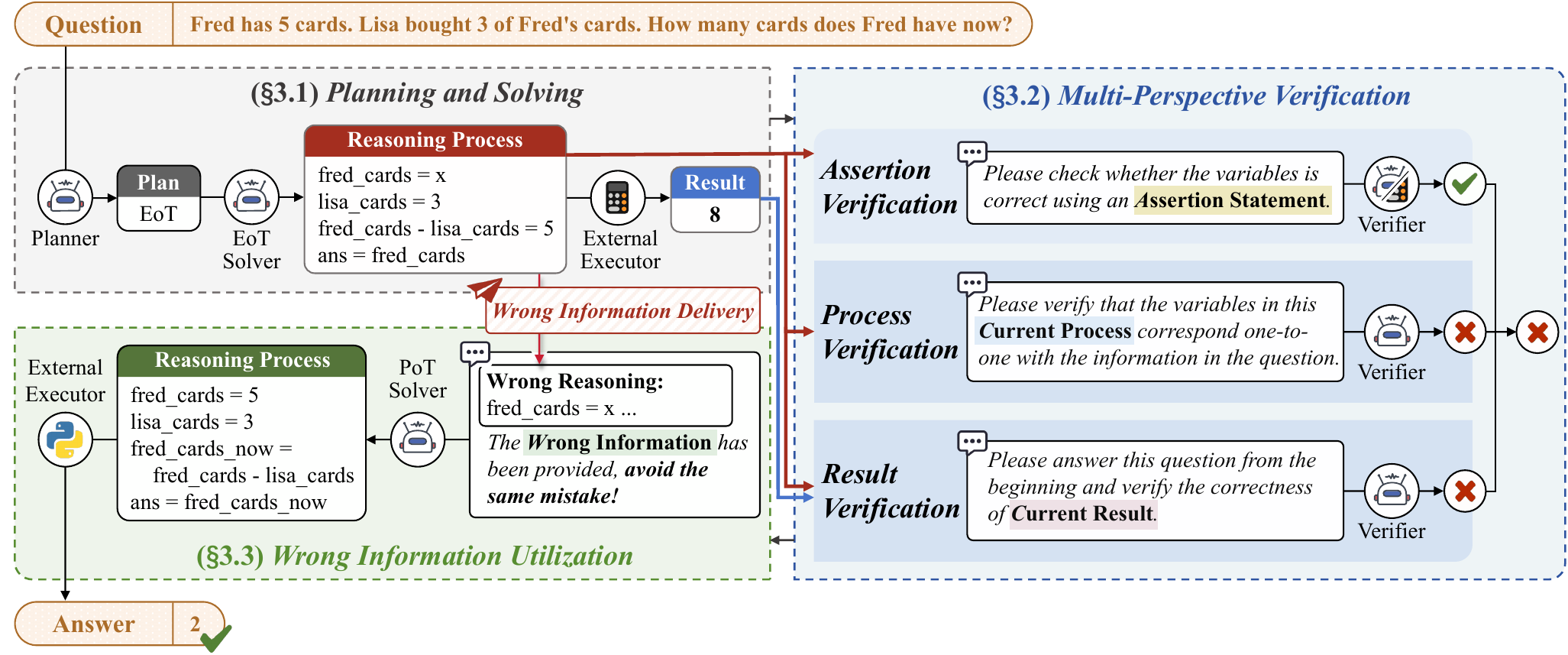}
	\caption{Overview of the Wrong-of-Thought (\modelname{}) framework, incorporating three core modules: \textit{Planning and Solving} ($\S \ref{plan}$), \textit{\textbf{\Blue{Multi-Perspective Verification}}} ($\S \ref{MPV}$), and \textit{\textbf{\Green{Wrong Information Utilization}}} ($\S \ref{WRA}$).}
	\label{framework}
\end{figure*}

This section introduces the framework that mainstream integrated multiple reasoning thoughts, iteratively enhancing LLMs by continuously re-verifying and refining corresponding reasoning. XoT \cite{liu2023plan}, as shown in Figure~\ref{back}, is an integrated reasoning framework that combines three reasoning modes: Program-of-Thought (PoT) \cite{chen2022program}, Equation-of-Thought (EoT) \cite{liu2023plan}, and Chain-of-Thought (CoT) \cite{wei2022chain}. PoT enables LLMs to generate Python code and then uses the external Python executor to run the results. EoT involves LLMs generating mathematical equations, which are then solved using an external calculator. CoT is a technique that guides LLMs to reason step-by-step.

In the XoT framework, the \textbf{First} step involves initiating the reasoning plan and selecting a reasoning method from either EoT or PoT to perform the reasoning. Once the reasoning process is completed, the result is computed through an external executor. The answer is then verified using an assertion verification.
If the reasoning result is determined to be correct, the answer is returned. If the initial reasoning is deemed incorrect and abandoned, the \textbf{Second} step is to switch to an alternative reasoning mode and restart the process. After obtaining and verifying the new reasoning answer, if it is still incorrect, the \textbf{Third} step is to directly use CoT reasoning as the final answer.

\section{Wrong-of-Thought}

\vspace{-0.1cm}

This section introduces Wrong-of-Thought (\modelname{}). The content is divided into three parts: \textit{Planning and Solving} ($\S \ref{plan}$), \textit{Multi-Perspective Verification} ($\S \ref{MPV}$), and \textit{Wrong Information Utilization} ($\S \ref{WRA}$).

\vspace{-0.1cm}

\subsection{Planning and Solving}
\label{plan}

Following XoT \cite{liu2023plan}, as shown in Figure~\ref{framework}~(§3.1), initially, a planner selects a reasoning method from either EoT or PoT based on the inputted question. After the Solver module generates the reasoning process, an external executor computes the result, yielding a preliminary solution. The next step is to validate the current solution.

\vspace{-0.1cm}

\subsection{Multi-Perspective Verification}
\label{MPV}

To address the challenge of verification methods being singular and significantly hindering the overall performance, we propose a \textit{Multi-Perspective Verification} (MPV), as shown in Figure~\ref{framework}~(§3.2). Specifically, \textit{Multi-Perspective Verification} is applicable to the reasoning verification of EoT and PoT, which includes the following three aspects:
\vspace{-0.05cm}
\begin{itemize}
\item [(1)]
\textbf{Assertion Verification}:
We adopt the verification method from the XoT \cite{liu2023plan}. We use LLMs to identify the intermediate variables in the solution and format them as \assertion{Assertion Statements}. These assertion statements are then executed using external tools to obtain the verification results.
\vspace{-0.1cm}
\item [(2)]
\textbf{Process Verification}:
For process verification, we provide the LLMs only with the \process{Current Process}, excluding the computed results. We ask the LLMs to recheck each step of the current reasoning process to ensure that the variables in the solution equations or code correspond one-to-one with the question information, explicitly demonstrating the verification reasoning process. 
\vspace{-0.1cm}
\item [(3)]
\textbf{Result Verification}:
In the results verification phase, we provide the LLMs with both the current reasoning process and the computed results. We instruct the LLMs to recheck the \result{Current Result} by re-solving the problem. If the result passes re-verification, the LLMs output ``right''; otherwise, they output ``error''. This explicitly demonstrates the verification reasoning results.
\end{itemize}

To enhance the robustness of our verification, we employ a voting mechanism to select the judgments that exhibit higher consistency across different verification perspectives $V_i$. These consistent judgments are then used as the final MVP results $\hat{V}$ for the output $R$ of the reasoning method $M_i$.
The verification can be formalized as follows:
\begin{equation}
\hat{V} = \underset{V_t \in \mathcal{V}}{\operatorname{argmax}}  \sum_{t=1}^{N} \sum_{R \in M_i} \mathds{1}(V_t = R),
\end{equation}
where $V_t$ represents verification methods, $\mathcal{V}$ represents the set of the three verification methods, $R$ represents the output using the reasoning method $M_i$, and $\mathds{1}(V_t = R)$ returns 1 if the verification method $V_t$ matches output $R$, and 0 otherwise.

\begin{table*}[t]
	\centering
	\begin{adjustbox}{width=0.98\textwidth}
	    \begin{tabular}{lccccccccc}
	    \toprule
	      Method    & GSM-hard & GSM8K & Algebra & MultiArith & SingleEQ & SingleOP & AddSub & SVAMP & Average \\
	    \midrule
	    \rowcolor[rgb]{ .970,  .970,  .970} 
	    \multicolumn{10}{c}{\textit{Mistral-7B-Instruct} \cite{jiang2023mistral}} \\
	    \midrule
	    \texttt{CoT} \cite{wei2022chain}  & 16.6  & 47.5  & 36.0  & 68.8  & 78.3  & 81.1  & 73.9  & 60.8  & 57.9  \\
	    \texttt{PoT} \cite{chen2022program}  & 30.8  & 45.0  & 28.4  & 72.8  & 75.8  & 64.4  & 74.7  & 56.5  & 56.0  \\
	    \texttt{EoT}  \cite{liu2023plan} & 16.1  & 22.3  & 27.0  & 25.0  & 31.1  & 33.6  & 29.1  & 23.5  & 26.0  \\
	    \texttt{XoT} \cite{liu2023plan}  & 26.2  & 52.8  & 46.8  & 77.8  & 86.6  & 85.4  & 80.0  & 67.9  & 65.5  \\
	    Wrong-of-Thought   & \textbf{36.7 } & \textbf{54.6 } & \textbf{50.5 } & \textbf{80.8 } & \textbf{88.0 } & \textbf{87.9 } & \textbf{88.9 } & \textbf{70.0 } & \textbf{69.7 } \\
	    \midrule
	    \rowcolor[rgb]{ .970,  .970,  .970} 
	    \multicolumn{10}{c}{\textit{Qwen-7B-Chat} \cite{qwen}} \\
	    \midrule
	    \texttt{CoT} \cite{wei2022chain}  & 18.6  & 52.8  & 43.7  & 83.2  & 87.4  & 83.1  & 80.5  & 70.7  & 65.0  \\
	    \texttt{PoT} \cite{chen2022program}  & 39.0  & 56.2  & 38.7  & 84.8  & 90.6  & 89.7  & 82.5  & 71.3  & 69.1  \\
	    \texttt{EoT} \cite{liu2023plan}  & 35.3  & 49.2  & 34.2  & 61.5  & 76.0  & 63.5  & 65.1  & 48.0  & 54.1  \\
	    \texttt{XoT} \cite{liu2023plan}  & 38.3  & 61.8  & 54.5  & 88.7  & 92.1  & 92.3  & 85.1  & 76.4  & 73.6  \\
	    Wrong-of-Thought   & \textbf{42.0 } & \textbf{63.7 } & \textbf{57.2 } & \textbf{91.3 } & \textbf{94.1 } & \textbf{93.6 } & \textbf{86.3 } & \textbf{79.3 } & \textbf{75.9 } \\
	    \midrule
	    \rowcolor[rgb]{ .970,  .970,  .970} \multicolumn{10}{c}{\textit{Qwen-14B-Chat} \cite{qwen}} \\
	    \midrule
	    \texttt{CoT} \cite{wei2022chain}  & 31.0  & 63.4  & 56.8  & 89.8  & 88.0  & 85.4  & 85.3  & 80.8  & 72.6  \\
	    \texttt{PoT} \cite{chen2022program}  & 57.1  & 69.5  & 62.6  & 95.7  & 95.7  & 96.1  & 86.8  & 81.6  & 80.6  \\
	    \texttt{EoT} \cite{liu2023plan}  & 57.6  & 68.5  & 62.6  & 85.7  & 90.6  & 82.2  & 83.8  & 79.2  & 76.3  \\
	    \texttt{XoT} \cite{liu2023plan}  & 55.3  & 76.3  & 80.2  & 92.0  & 94.1  & 94.5  & 86.1  & 84.8  & 82.9  \\
	    Wrong-of-Thought   & \textbf{60.6 } & \textbf{77.5 } & \textbf{81.5 } & \textbf{98.3 } & \textbf{96.7 } & \textbf{95.4 } & \textbf{88.1 } & \textbf{86.3 } & \textbf{85.5 } \\
	    \midrule
	    \rowcolor[rgb]{ .970,  .970,  .970} \multicolumn{10}{c}{\textit{Gemini-1.0-Pro} \cite{team2023gemini}} \\
	    \midrule
	    \texttt{CoT} \cite{wei2022chain}  & 45.6  & 81.9  & 81.5  & 94.8  & 96.1  & 94.7  & 92.9  & 83.0  & 83.8  \\
	    \texttt{PoT} \cite{chen2022program}  & 63.8  & 77.1  & 58.1  & 96.3  & 96.3  & 96.3  & 91.6  & 87.1  & 83.3  \\
	    \texttt{EoT} \cite{liu2023plan}  & 52.2  & 61.1  & 63.5  & 80.0  & 79.7  & 75.3  & 78.0  & 71.3  & 70.1  \\
	    \texttt{XoT}  \cite{liu2023plan} & 64.6  & 82.1  & 83.3  & 96.5  & 96.1  & 96.3  & 91.4  & 86.9  & 87.2  \\
	    Wrong-of-Thought   & \textbf{69.1 } & \textbf{84.4 } & \textbf{85.6 } & \textbf{97.3 } & \textbf{97.4 } & \textbf{97.3 } & \textbf{93.4 } & \textbf{89.2 } & \textbf{89.2 } \\
	    \midrule
	    \rowcolor[rgb]{ .970,  .970,  .970} \multicolumn{10}{c}{\textit{GPT-3.5-Turbo} \cite{openai2022chatgpt}} \\  
	    \midrule
	    \texttt{CoT} \cite{wei2022chain} & 42.2  & 80.0  & 72.1  & 97.3  & 96.5  & 94.7  & 89.4  & 80.2  & 81.5  \\
	    \texttt{PoT} \cite{chen2022program}  & 70.3  & 77.4  & 81.5  & 97.8  & 98.6  & 94.3  & 88.9  & 79.2  & 86.0  \\
	    \texttt{EoT} \cite{liu2023plan}  & 53.4  & 64.0  & 70.3  & 84.8  & 61.4  & 68.5  & 70.1  & 58.9  & 66.4  \\
	    \texttt{XoT} \cite{liu2023plan}  & 71.3  & 83.6  & 84.7  & 97.8  & 97.6  & 94.5  & 89.4  & 83.0  & 87.7  \\ 
	    Wrong-of-Thought   & \textbf{76.2 } & \textbf{85.2 } & \textbf{89.6 } & \textbf{99.0 } & \textbf{99.0 } & \textbf{96.1 } & \textbf{93.2 } & \textbf{86.7 } & \textbf{90.6 } \\
	    \bottomrule
	    \end{tabular}
  \end{adjustbox}
  	\caption{Experimental results of Acc. (\%) on eight datasets and five LLMs. \textbf{Bold} represents the best performance.}
	\label{main results}
\end{table*}
	
\subsection{Wrong Information Utilization}
\label{WRA}

To address the issue of previous methods ignoring wrong information, we propose \textit{Wrong Information Utilization} (WIU), as shown in Figure~\ref{framework}~(§3.3).
Specifically, after the previous solution is validated and determined to be wrong, we incorporate the prior \wrong{Wrong Information} within the context of the current solution method. This guides the LLMs to avoid repeating the same mistakes. Formally, the reasoning for the question $Q$ after utilizing wrong reasoning information can be expressed by the following formula:
\begin{equation}
	\hat{R} = \underset{ R \in M_i}{\operatorname{argmax}} \ P(R|Q, I, \textrm{WI}),
\end{equation}
where $\hat{R}$ represents the final reasoning result. $P(R|Q, I, \textrm{WI})$ denotes the probability of generating the reasoning path $R$ under the conditions of question $Q$, prompt $I$, and \wrong{Wrong Information} $\textrm{WI}$. $R$ is a reasoning of the reasoning method $M_i$.

After obtaining the reasoning results, we use the \textit{Multi-Perspective Verification} to make a judgment. If the judgment is correct, the answer is returned directly. If the judgment is wrong, following XoT, we proceed to the third step, where the errors from this step and the previous step will be used as wrong examples for CoT reasoning.
	
\section{Experiments}

\subsection{Experimental Setting}

We conduct experiments on eight widely used comprehensive datasets, including GSM8K \cite{cobbe2021training}, GSM-Hard \cite{gao2023pal}, Algebra \cite{he2023solving}, MultiArith \cite{roy-roth-2015-solving}, SingleEQ \cite{Koncel-Kedziorski_Hajishirzi_Sabharwal_Etzioni_Ang_2015}, SingleOP \cite{roy2015reasoning}, AddSub \cite{Hosseini_Hajishirzi_Etzioni_Kushman_2014}, and SVAMP \cite{patel-etal-2021-nlp}. The effectiveness of the \modelname{} framework was validated on these challenging benchmarks. 

Additionally, we select the single reasoning methods CoT \cite{wei2022chain}, PoT \cite{chen2022program}, and EoT \cite{liu2023plan}, as well as the ensemble method XoT \cite{liu2023plan}, as baselines. The verification process was conducted on a comprehensive set of five LLMs. Among these, three are open-source LLMs: Mistral-7B-Instruct \cite{jiang2023mistral}, Qwen-7B-Chat \cite{qwen}, and Qwen-14B-Chat \cite{qwen}. The other two LLMs are closed-source: Gemini-1.0-Pro \cite{team2023gemini} and GPT-3.5-Turbo \cite{openai2022chatgpt}. These models were selected to provide a diverse representation of current advanced LLMs, both open and closed-source, ensuring a robust and comprehensive verification.

Following XoT \cite{liu2023plan}, all experiments used 8-shot correct examples as prompts. The experimental results were evaluated using Accuracy as the evaluation metric. The top-p and temperature parameters for all experiments were set to LLMs default parameters in the official model configuration, which are within the range of [0,1].

\subsection{Main Results}

The main experimental results are shown in Table~\ref{main results}. Based on the results, we can observe:
\vspace{0.5\baselineskip}

\noindent
\textbf{(1) \modelname{} reaches superior performance.}
\modelname{} surpasses all baselines, achieving superior performance on eight datasets, with an average improvement of 2.8\% compared to XoT across five LLMs. This extensive experimental result demonstrates the effectiveness of the integration of \textit{Multi-Perspective Verification} and \textit{Wrong Information Utilization} in \modelname{}, enhancing overall performance.

\vspace{0.5\baselineskip}
\noindent
\textbf{(2) \modelname{} can also work on LLMs with smaller parameters.}
\modelname{} achieves an average improvement of 4.2\% and 2.3\% on the smaller parameter open-source models, Mistral-7B-Instruct and Qwen1.5-7B-Chat, respectively, demonstrating robust performance. The ability of \modelname{} to maintain high performance on models with fewer parameters highlights its potential for broad applicability in various practical scenarios, including those with limited computational resources.

\vspace{0.5\baselineskip}
\noindent
\textbf{(3) \modelname{} demonstrates a powerful ability to solve difficult reasoning questions. }
\modelname{} achieves an average performance on GSM-Hard that was 5.7\% higher than the baselines on five LLMs, representing a significant improvement. The GSM-Hard dataset, a mathematical reasoning dataset where small numerical values are replaced with large ones (average result: $7.3e9$), demonstrates the strong performance of \modelname{} in difficult reasoning tasks.

\subsection{\modelname{} Analysis}

To gain a more profound understanding of \modelname{}, we propose the following research questions based on experiments on GPT-3.5-Turbo \cite{openai2022chatgpt}:

\begin{table*}
	\centering
	\begin{adjustbox}{width=\textwidth}
		\begin{tabular}{lccccccccl}
			\toprule
			Methods & GSM-hard & GSM8K & Algebra & MultiArith & SingleEQ & SingleOP & AddSub & SVAMP & AVG  \\
			\midrule
			Wrong-of-Though & 76.2  & 85.2  & 89.6 & 99.0  & 99.0  & 96.1  & 93.2 & 86.7 & 90.6 \\
			\midrule
			\ \ \textit{w/o WIU} & 73.9 & 84.0 & 87.8 & 98.8 & 98.4 & 95.9 & 92.6 & 85.5  & 89.6 \textit{\textcolor{red}{(-1.0)}}\\
			\ \ \textit{w/o MPV} & 73.1 & 82.4 & 87.4 & 98.3 & 98.6 & 94.5 & 90.4 & 85.6 &  88.8 \textit{\textcolor{red}{(-1.8)}} \\
			\ \ \textit{w/o WIU \& MPV} &  71.3  & 83.6  & 84.7  & 97.8  & 97.6  & 94.5  & 89.4  & 83.0  & 87.7 \textit{\textcolor{red}{(-2.9)}} \\ 
			\bottomrule
			\end{tabular}
	\end{adjustbox}
	\caption{Ablation experiment on GPT-3.5-Turbo. ``\textit{w/o WIU}'' refers to removing \textit{Wrong Information Utilization} (WIU). ``\textit{w/o MPV}'' refers to removing \textit{Multi-Perspective Verification} (MPV). ``\textit{w/o WIU \& MPV}'' refers to removing both \textit{Wrong Information Utilization} and \textit{Multi-Perspective Verification}. }
	\label{ablation}
\end{table*}

\begin{figure*}[t]
	\centering
	\includegraphics[width=160mm]{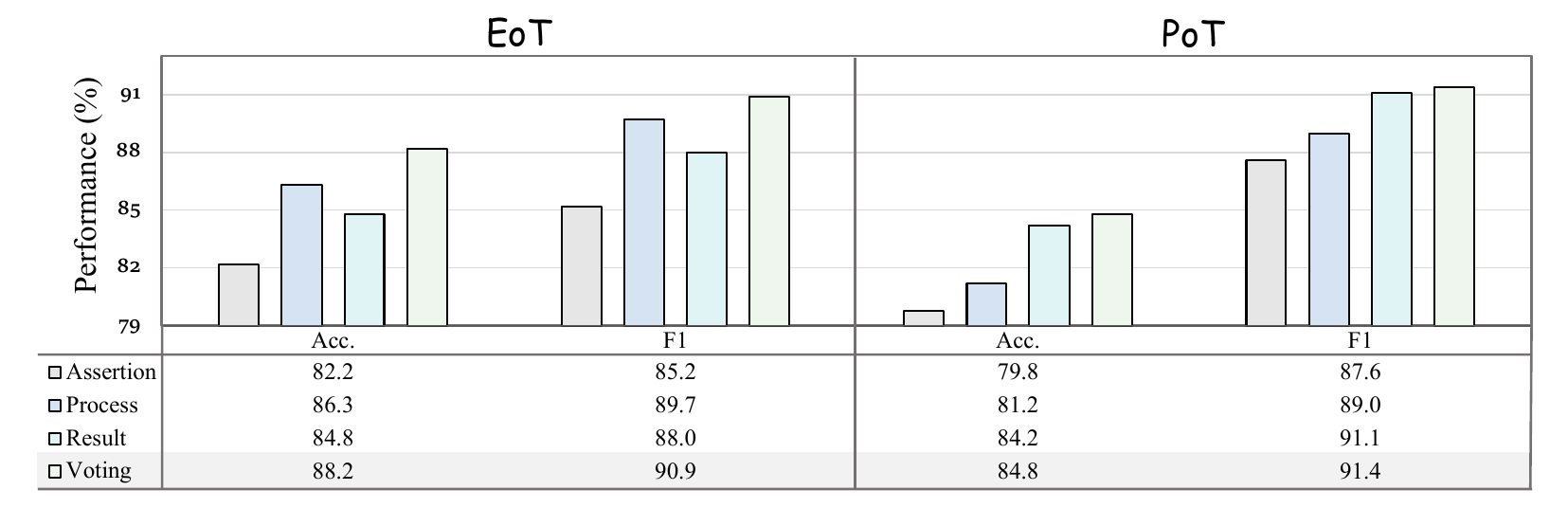}
	\caption{Performance comparison results from various verification perspectives. ``Voting'' represents the final judgment after voting from the three perspectives.}
	\label{verif}
\end{figure*}

\begin{itemize}
\item [(1)]
\textit{Can Wrong Information Utilization lead to performance improvement?}

\item [(2)] \textit{Can Multi-Perspective Verification lead to more accurate judgment results?}

\item [(3)] \textit{Can \modelname{} reduce the number of required reasoning steps?}

\item [(4)] \textit{Why does \modelname{} have strong capabilities in difficult mathematical reasoning?}

\item [(5)] \textit{What is the intuition behind \modelname{}?}
\end{itemize}

\subsubsection{Answer 1: \textit{Wrong Information Utilization} can boost performance}

To intuitively verify the performance improvements brought by using wrong information, we select PoT, EoT, and CoT that utilized wrong information from the GSM8K dataset for evaluation.
We compare their performance with and without wrong information. Additionally, we test the \modelname{} performance without the \textit{Wrong Information Utilization}. 
Due to the limitation within the \modelname{}, EoT and PoT can only collect incorrect information once, resulting in a single wrong example. On the other hand, CoT can collect incorrect information up to two times, resulting in two wrong examples.

\begin{figure}[t]
	\centering
	\includegraphics[width=0.49\textwidth]{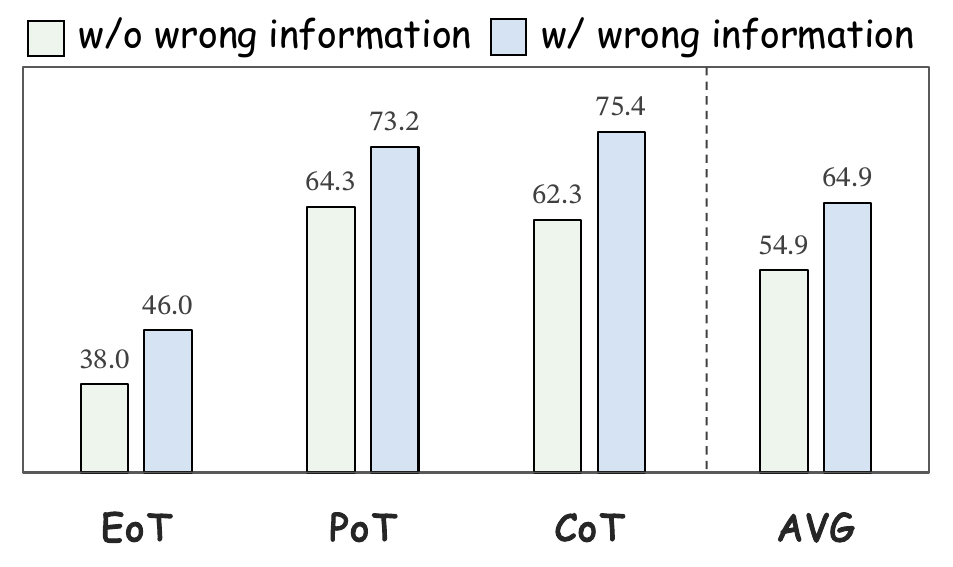}
	\caption{Comparison of performance without utilizing wrong reasoning information and with integrated wrong reasoning information.}
	\label{wrong_info}
\end{figure}

The results are shown in the Figure~\ref{wrong_info}. After incorporating wrong information from the previous step, EoT and PoT improved by 8\% and 8.9\%, respectively. We can observe that CoT, which utilized additional wrong information from the previous two steps, improved by 13.1\%. 
Furthermore, as shown in Table~\ref{ablation}, the \modelname{} framework without \textit{Wrong Information Utilization} exhibits a performance decrease across all datasets, with an average reduction of 1.0\%.
This demonstrates that incorporating wrong information can boost the reasoning performance of the LLMs, and more significant improvements can be achieved by utilizing more additional wrong reasoning information.

\subsubsection{Answer 2: \textit{Multi-Perspective Verification} can lead to more accurate judgments}

To demonstrate that \textit{Multi-Perspective Verification} can accurately judge the results generated by EoT and PoT, we directly evaluated the performance of the three perspectives and the final voting results of the three perspectives. For accurate assessment, we use accuracy (Acc.) and F1 score (F1) as evaluation metrics. Additionally, we evaluate the performance of the \modelname{} framework without \textit{Multi-Perspective Verification} to demonstrate the effectiveness of \textit{Multi-Perspective Verification}.

The results are shown in Figure~\ref{verif}. 
We can directly observe that our proposed Process Verification and Result Verification outperform the Assertion Verification used in XoT with respect to accuracy and F1 score. Furthermore, the final Voting Verification further improves the accuracy. For EoT, Acc and F1 improved by 6\% and 5.7\%, respectively, while for PoT, they improved by 5\% and 3.8\%, respectively. 
Additionally, as shown in Table~\ref{ablation}, the performance of \modelname{} decreased by an average of 1.8\% after the removal of \textit{Multi-Perspective Verification}.
This demonstrates the effectiveness of \textit{Multi-Perspective Verification}, bringing significant benefits to overall performance improvement.

\begin{figure}[t]
	\centering
	\includegraphics[width=0.54\textwidth]{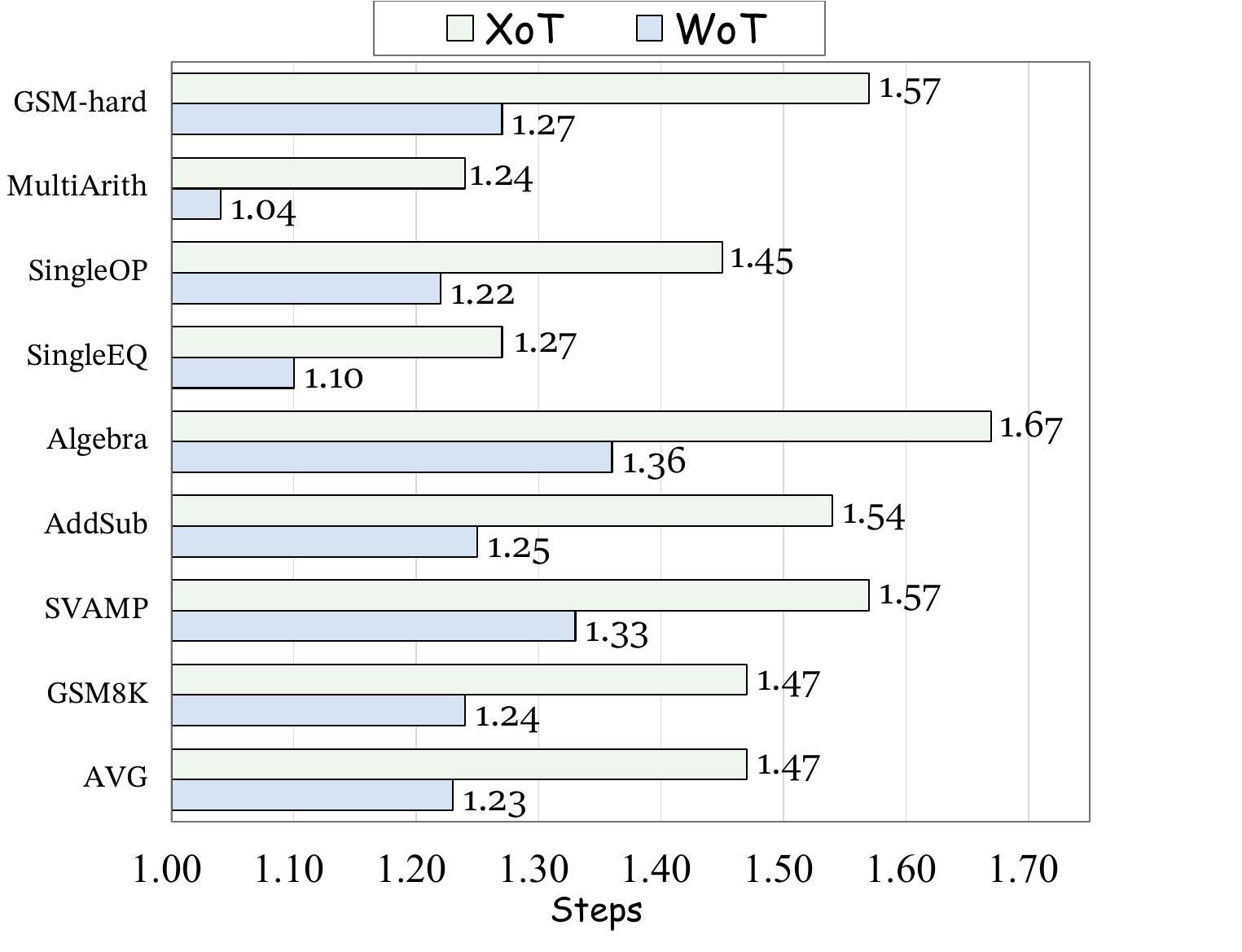}
	\caption{Comparison of the average reasoning steps required by XoT and \modelname{} in solving questions.}
	\label{step}
\end{figure}

\subsubsection{Answer 3: \modelname{} can effectively minimize the reasoning steps needed}

To compare the reasoning steps required by XoT and \modelname{} in solving mathematical questions, we conduct experiments and record the average reasoning steps needed. As shown in Figure~\ref{step}, the results indicate that \modelname{} significantly reduces the reasoning steps in each dataset, with an average reduction of 8\% steps. This indirectly demonstrates the effectiveness of \textit{Multi-Perspective Verification}, and \textit{Wrong Information Utilization} in \modelname{}. Accurate verification and efficient reasoning can effectively reduce the number of required reasoning steps, thereby enhancing reasoning efficiency.

\subsubsection{Answer 4: Tips for solving difficult mathematical questions with \modelname{}}

\begin{figure}[t]
	\centering
	\includegraphics[width=0.49\textwidth]{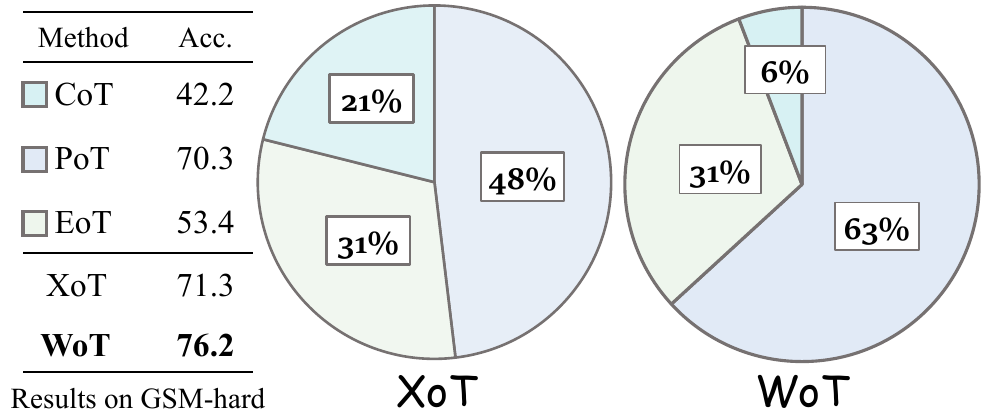}
	\caption{The proportion of reasoning methods ultimately used to solve questions by XoT and \modelname{} on the GSM-hard dataset.}
	\vspace{-0.5\baselineskip}
	\label{ratio}
\end{figure}

To delve deeper into the reasons behind the significant performance improvement of \modelname{} in solving reasoning challenges, we conduct a detailed analysis in this section. In the GSM-hard dataset, we extract the proportions of the methods ultimately used for reasoning, as shown in Figure~\ref{ratio}. Our analysis reveals notable changes in the reasoning method proportions between XoT and \modelname{}: the proportion of CoT decreased from 21\%~$\rightarrow$~6\%, while the proportion of PoT increased from 48\%~$\rightarrow$~63\%.

This change reflects the advantage of \modelname{} in reasoning strategies. The numerical values in the GSM-hard dataset are usually large, often involving more than 10 digits. Because CoT reasoning has lower accuracy when handling large number value calculations, with an accuracy rate of only 42.2\%. Since XoT relies more on CoT for reasoning, it results in lower accuracy. In contrast, \modelname{} introduces a multiple perspectives verification mechanism, enabling more accurate judgment of reasoning results. Consequently, \modelname{} more frequently adopts PoT for reasoning, thereby avoiding errors associated with CoT, and achieving significant overall improvement.

\begin{figure*}[t]
	\centering
	\includegraphics[width=160mm]{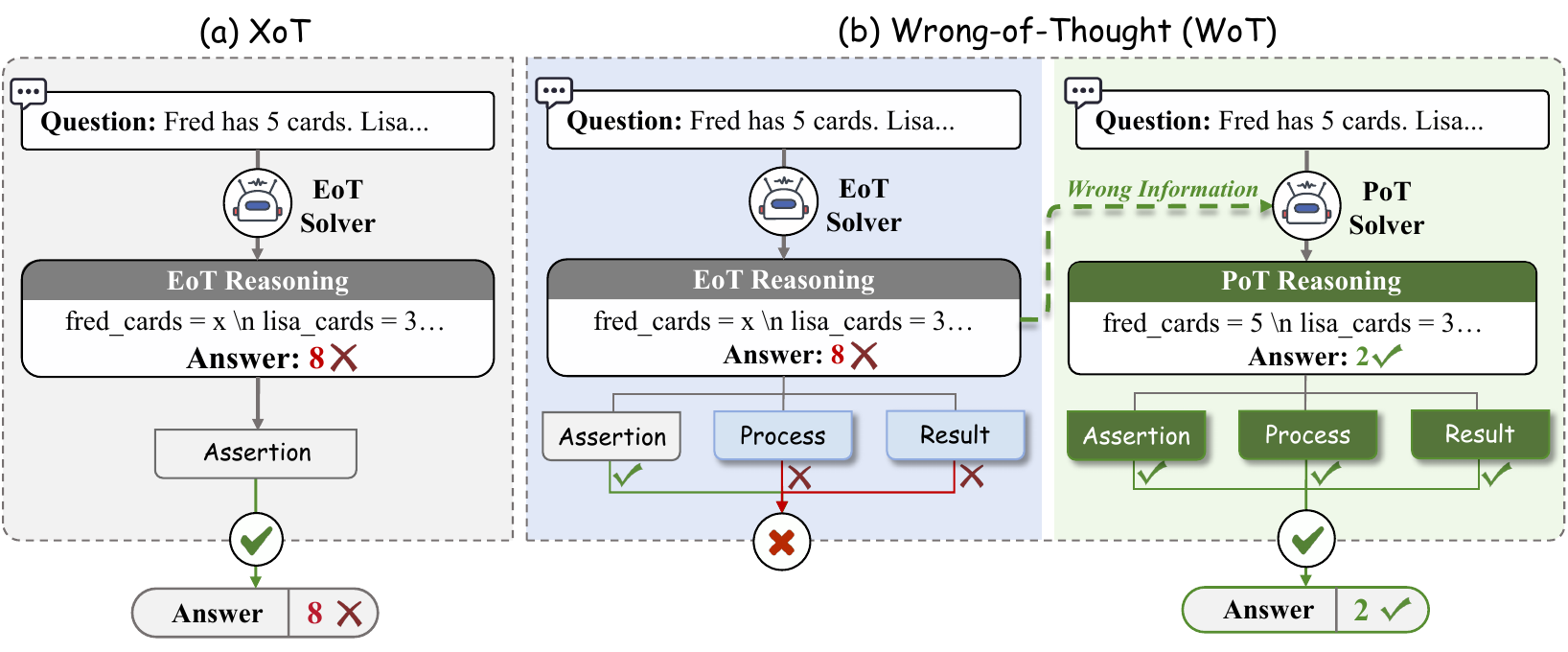}
	\caption{The case study. Figure (a) shows XoT reasoning, where it initially outputs an incorrect answer, ``8''. Assertion Verification mistakenly validated this as correct, resulting in the final wrong output of ``8''. Figure (b) shows \modelname{} reasoning. EoT first outputs an incorrect answer, which was identified as wrong by Process and Result Verification, switching to PoT. Using the wrong reasoning of EoT, PoT arrived at the correct answer, ``2''. All three verification methods then confirmed this result, leading to the correct output of ``2''.}
	\label{case}
\end{figure*}

\subsubsection{Answer 5: Qualitative analysis}

To better comprehend \modelname{}, we introduce a real-world example for qualitative analysis. As illustrated in Figure~\ref{case}~(a), upon receiving a question, XoT selects EoT for reasoning. However, due to the limited reasoning capability of EoT, an incorrect result of ``8'' was generated. During Assertion Verification, this incorrect result was mistakenly identified as correct. As XoT relied solely on Assertion Verification, it erroneously output ``8'' as the final result. This example clearly illustrates the limitations of the single verification method and its adverse impact on reasoning accuracy.

In contrast, as shown in Figure~\ref{case}~(b), \modelname{}, when presented with the same question, initially also arrives at the incorrect answer ``8''. However, both Process Verification and Result Verification identified ``8'' as incorrect. Consequently, the system switches to PoT for the next reasoning step. In PoT reasoning, after being warned with a wrong example, PoT generates the correct reasoning and arrives at the correct result, ``2''. This result then passed verification from all three perspectives, ultimately confirming the correct answer, ``2''. This case further demonstrates the effectiveness of \modelname{}, as combining three verification perspectives and utilizing wrong reasoning information significantly enhances reasoning capability.
	
\section{Related Work}

The rapid advancement of LLMs in recent years has introduced new opportunities in natural language processing \cite{openai2022chatgpt, team2023gemini,qin2024large,qin2024multilingual}. Particularly, the introduction of Chain-of-Thought (CoT) \cite{wei2022chain} opens a novel direction in this domain, attracting many researchers \cite{zhang2022automatic,fei2023reasoning,fei2024video,zhang2024autocap,xu2024faithful,chen2024m}. Specifically, \citet{wei2022chain} propose using manually constructed CoT demonstrations to enhance LLMs performance. Additionally, \citet{chen2022program} introduce the Program-of-Thoughts (PoT), enabling LLMs to generate Python programs to solve mathematical problems. \citet{liu2023plan} propose Equation-of-Thoughts (EoT), allowing LLMs to generate mathematical equations and then use external calculators to compute the results, offering a new perspective on problem-solving with LLMs. \citet{chen2024boosting} propose a framework that iteratively explores and self-evaluates trees of thoughts, allowing LLMs to learn from trial and error and improve the quality of final answers.  \citet{xu2024activerag} propose transitioning LLMs from passive to active learning, thus enhancing their problem-solving capabilities. \citet{zhou2024self} present a method for LLMs to improve self-criticism and self-discovery, thereby forming explicit structures to enhance reasoning performance. \citet{chen2023teaching} propose using error code to implement code self-debug and improve the code generation capability of LLMs.
\vspace{-0.03\baselineskip}

In the realm of nonlinear problem solving, \citet{yao2024tree} introduce the Tree-of-Thoughts (ToT) framework, enabling LLMs to generate multiple reasoning paths to tackle mathematical reasoning tasks. \citet{sel2023algorithm} propose the Algorithm-of-Thoughts (AoT), which not only generates multiple paths but also selects the optimal nodes, allowing for the repeated utilization of reasoning pathways. \citet{besta2024graph} introduce Graph-of-Thoughts (GoT), a framework that models the information generated by LLMs as arbitrary graphs, enabling the synergistic integration of all reasoning processes. \citet{ning2024skeleton} propose Skeleton-of-Thought (SoT), which first generates the skeleton of the answer and then utilizes LLMs for batched resolution, enhancing inference efficiency.
\citet{liu2023plan} propose XoT, which integrates multiple reasoning thoughts and utilizes single assertion verification to decide whether to switch reasoning methods, achieving impressive results.

Compared to previous research, \modelname{} employs multiple perspectives of verification while incorporating wrong information utilization. This greatly and effectively enhances overall reasoning performance. To our knowledge, this work is the first to incorporate \textit{Multi-Perspective Verification} and \textit{Wrong Information Utilization} within the continuously verifying and iterative framework.
	
\section{Conclusion}

In this work, we propose \modelname{}, a framework that optimizes outputs by utilizing wrong information and multi-perspective verification. \modelname{} comprises two core modules: \textit{Multi-Perspective Verification} and \textit{Wrong Information Utilization}. \modelname{} achieves more accurate reasoning thought switching and utilizes wrong reasoning information. Extensive evaluations on eight datasets and five models demonstrate that \modelname{} achieves superior performance. Furthermore, \modelname{} exhibits powerful capabilities in difficult computation tasks.

\section*{Limitations}
This work proposes a \modelname{} framework to enhance verifying iteratively generated reasoning answers by \textit{Multi-Perspective Verification} and \textit{Wrong Information Utilization}.
However, in our work, since ``\textit{Assertion Verification}'' requires reliance on external rule executors, how to verify natural language-based CoT through assertions remains a question worthy of future research. Secondly, our verification method primarily validates the logical correctness of the model. Verifying the clarity and quality of the logical expression might further enhance the effectiveness of model reasoning. Finally, \modelname{} may spend more tokens due to the incorporation of three verification perspectives and wrong reasoning information. We hope future work develops more efficient methods to address this challenge.

\section*{Acknowledgments}
This work was supported by the National Natural Science Foundation of China (NSFC) via grant 62306342. This work was also sponsored by the Excellent Young Scientists Fund in Hunan Province (2024JJ4070) and the Science and Technology Innovation Program of Hunan Province under Grant 2024RC3024. This work was supported by the Key Laboratory of Computing Power Network and Information Security, Ministry of Education under Grant No.2023ZD032. We are grateful for resources from the High Performance Computing Center of Central South University.

\bibliography{custom}

\begin{thebibliography}{40}
\providecommand{\natexlab}[1]{#1}

\bibitem[{Achiam et~al.(2023)Achiam, Adler, and Agarwal}]{achiam2023gpt}
Josh Achiam, Steven Adler, and Sandhini et~al. Agarwal. 2023.
\newblock \href {https://arxiv.org/abs/2303.08774} {Gpt-4 technical report}.
\newblock abs/2303.08774.

\bibitem[{Bai et~al.(2023)Bai, Bai, and et~al.}]{qwen}
Jinze Bai, Shuai Bai, and Yunfei~Chu et~al. 2023.
\newblock \href {https://arxiv.org/abs/2309.16609} {Qwen technical report}.
\newblock \emph{ArXiv preprint}, abs/2309.16609.

\bibitem[{Besta et~al.(2024)Besta, Blach, and Kubicek}]{besta2024graph}
Maciej Besta, Nils Blach, and Ales et~al. Kubicek. 2024.
\newblock Graph of thoughts: Solving elaborate problems with large language models.
\newblock In \emph{Proceedings of the AAAI Conference on Artificial Intelligence}, volume~38, pages 17682--17690.

\bibitem[{Chen et~al.(2024{\natexlab{a}})Chen, Qin, Zhang, Chen, Xu, and Che}]{chen2024m}
Qiguang Chen, Libo Qin, Jin Zhang, Zhi Chen, Xiao Xu, and Wanxiang Che. 2024{\natexlab{a}}.
\newblock \href {https://doi.org/10.18653/v1/2024.acl-long.446} {{M}$^3${C}o{T}: A novel benchmark for multi-domain multi-step multi-modal chain-of-thought}.
\newblock In \emph{Proceedings of the 62nd Annual Meeting of the Association for Computational Linguistics (Volume 1: Long Papers)}, pages 8199--8221.

\bibitem[{Chen et~al.(2024{\natexlab{b}})Chen, Li, and Niu}]{chen2024boosting}
Sijia Chen, Baochun Li, and Di~Niu. 2024{\natexlab{b}}.
\newblock \href {https://openreview.net/forum?id=qBL04XXex6} {Boosting of thoughts: Trial-and-error problem solving with large language models}.
\newblock In \emph{The Twelfth International Conference on Learning Representations}.

\bibitem[{Chen et~al.(2022)Chen, Ma, Wang, and Cohen}]{chen2022program}
Wenhu Chen, Xueguang Ma, Xinyi Wang, and William~W Cohen. 2022.
\newblock Program of thoughts prompting: Disentangling computation from reasoning for numerical reasoning tasks.
\newblock \emph{Transactions on Machine Learning Research}.

\bibitem[{Chen et~al.(2023)Chen, Lin, Schaerli, and Zhou}]{chen2023teaching}
Xinyun Chen, Maxwell Lin, Nathanael Schaerli, and Denny Zhou. 2023.
\newblock Teaching large language models to self-debug.
\newblock In \emph{The 61st Annual Meeting Of The Association For Computational Linguistics}.

\bibitem[{Cobbe et~al.(2021)Cobbe, Kosaraju, Bavarian, Chen, Jun, Kaiser, Plappert, Tworek, Hilton, Nakano et~al.}]{cobbe2021training}
Karl Cobbe, Vineet Kosaraju, Mohammad Bavarian, Mark Chen, Heewoo Jun, Lukasz Kaiser, Matthias Plappert, Jerry Tworek, Jacob Hilton, Reiichiro Nakano, et~al. 2021.
\newblock \href {https://arxiv.org/abs/2110.14168} {Training verifiers to solve math word problems}.
\newblock \emph{ArXiv preprint}, abs/2110.14168.

\bibitem[{Fei et~al.(2023)Fei, Li, Liu, Bing, Li, and Chua}]{fei2023reasoning}
Hao Fei, Bobo Li, Qian Liu, Lidong Bing, Fei Li, and Tat-Seng Chua. 2023.
\newblock Reasoning implicit sentiment with chain-of-thought prompting.
\newblock In \emph{Proceedings of the 61st Annual Meeting of the Association for Computational Linguistics (Short Papers)}.

\bibitem[{Fei et~al.(2024)Fei, Wu, Ji, Zhang, Zhang, Lee, and Hsu}]{fei2024video}
Hao Fei, Shengqiong Wu, Wei Ji, Hanwang Zhang, Meishan Zhang, Mong-Li Lee, and Wynne Hsu. 2024.
\newblock Video-of-thought: Step-by-step video reasoning from perception to cognition.
\newblock In \emph{Forty-first International Conference on Machine Learning}.

\bibitem[{Gao et~al.(2023)Gao, Madaan, and Zhou}]{gao2023pal}
Luyu Gao, Aman Madaan, and Shuyan et~al. Zhou. 2023.
\newblock Pal: Program-aided language models.
\newblock In \emph{International Conference on Machine Learning}, pages 10764--10799. PMLR.

\bibitem[{He-Yueya et~al.(2023)He-Yueya, Poesia, Wang, and Goodman}]{he2023solving}
Joy He-Yueya, Gabriel Poesia, Rose~E Wang, and Noah~D Goodman. 2023.
\newblock \href {https://arxiv.org/abs/2304.09102} {Solving math word problems by combining language models with symbolic solvers}.
\newblock \emph{ArXiv preprint}, abs/2304.09102.

\bibitem[{Hosseini et~al.(2014)Hosseini, Hajishirzi, Etzioni, and Kushman}]{Hosseini_Hajishirzi_Etzioni_Kushman_2014}
Mohammad~Javad Hosseini, Hannaneh Hajishirzi, Oren Etzioni, and Nate Kushman. 2014.
\newblock \href {https://doi.org/10.3115/v1/D14-1058} {Learning to solve arithmetic word problems with verb categorization}.
\newblock In \emph{Proceedings of the 2014 Conference on Empirical Methods in Natural Language Processing ({EMNLP})}, pages 523--533.

\bibitem[{Jiang et~al.(2023)Jiang, Sablayrolles, and Mensch}]{jiang2023mistral}
Albert~Q Jiang, Alexandre Sablayrolles, and Arthur et~al. Mensch. 2023.
\newblock \href {https://arxiv.org/abs/2310.06825} {Mistral 7b}.
\newblock \emph{ArXiv preprint}, abs/2310.06825.

\bibitem[{Koncel-Kedziorski et~al.(2015)Koncel-Kedziorski, Hajishirzi, Sabharwal, Etzioni, and Ang}]{Koncel-Kedziorski_Hajishirzi_Sabharwal_Etzioni_Ang_2015}
Rik Koncel-Kedziorski, Hannaneh Hajishirzi, Ashish Sabharwal, Oren Etzioni, and Siena~Dumas Ang. 2015.
\newblock \href {https://doi.org/10.1162/tacl_a_00160} {Parsing algebraic word problems into equations}.
\newblock \emph{Transactions of the Association for Computational Linguistics}, 3:585--597.

\bibitem[{Liu et~al.(2023{\natexlab{a}})Liu, Teng, Cui, Zhang, Zhou, and Zhang}]{liu2023logicot}
Hanmeng Liu, Zhiyang Teng, Leyang Cui, Chaoli Zhang, Qiji Zhou, and Yue Zhang. 2023{\natexlab{a}}.
\newblock Logicot: Logical chain-of-thought instruction tuning.
\newblock In \emph{The 2023 Conference on Empirical Methods in Natural Language Processing}.

\bibitem[{Liu et~al.(2023{\natexlab{b}})Liu, Guo, Yang, Hu, Zhang, Qiu, and Zhang}]{liu2023plan}
Tengxiao Liu, Qipeng Guo, Yuqing Yang, Xiangkun Hu, Yue Zhang, Xipeng Qiu, and Zheng Zhang. 2023{\natexlab{b}}.
\newblock \href {https://doi.org/10.18653/v1/2023.emnlp-main.169} {Plan, verify and switch: Integrated reasoning with diverse {X}-of-thoughts}.
\newblock In \emph{Proceedings of the 2023 Conference on Empirical Methods in Natural Language Processing}, pages 2807--2822.

\bibitem[{Madaan et~al.(2023)Madaan, Tandon, and Gupta}]{madaan2023self}
Aman Madaan, Niket Tandon, and Prakhar et~al. Gupta. 2023.
\newblock Self-refine: Iterative refinement with self-feedback.
\newblock \emph{Advances in Neural Information Processing Systems}, 36.

\bibitem[{Ning et~al.(2024)Ning, Lin, and Zhou}]{ning2024skeleton}
Xuefei Ning, Zinan Lin, and Zixuan et~al. Zhou. 2024.
\newblock Skeleton-of-thought: Prompting llms for efficient parallel generation.
\newblock In \emph{The Twelfth International Conference on Learning Representations}.

\bibitem[{OpenAI(2022)}]{openai2022chatgpt}
OpenAI. 2022.
\newblock \href {https://platform.openai.com/docs/guides/text-generation/chat-completions-api} {Chatgpt}.

\bibitem[{Patel et~al.(2021)Patel, Bhattamishra, and Goyal}]{patel-etal-2021-nlp}
Arkil Patel, Satwik Bhattamishra, and Navin Goyal. 2021.
\newblock \href {https://doi.org/10.18653/v1/2021.naacl-main.168} {Are {NLP} models really able to solve simple math word problems?}
\newblock In \emph{Proceedings of the 2021 Conference of the North American Chapter of the Association for Computational Linguistics: Human Language Technologies}, pages 2080--2094.

\bibitem[{Qi et~al.(2023)Qi, Xu, Shen, and et~al.}]{qi2023art}
Jingyuan Qi, Zhiyang Xu, Ying Shen, and et~al. 2023.
\newblock The art of socratic questioning: Recursive thinking with large language models.
\newblock In \emph{Proceedings of the 2023 Conference on Empirical Methods in Natural Language Processing}, pages 4177--4199.

\bibitem[{Qin et~al.(2024{\natexlab{a}})Qin, Chen, Feng, Wu, Zhang, Li, Li, Che, and Yu}]{qin2024large}
Libo Qin, Qiguang Chen, Xiachong Feng, Yang Wu, Yongheng Zhang, Yinghui Li, Min Li, Wanxiang Che, and Philip~S Yu. 2024{\natexlab{a}}.
\newblock \href {https://arxiv.org/abs/2405.12819} {Large language models meet nlp: A survey}.
\newblock \emph{ArXiv preprint}, abs/2405.12819.

\bibitem[{Qin et~al.(2023)Qin, Chen, Wei, Huang, and Che}]{qin2023cross}
Libo Qin, Qiguang Chen, Fuxuan Wei, Shijue Huang, and Wanxiang Che. 2023.
\newblock Cross-lingual prompting: Improving zero-shot chain-of-thought reasoning across languages.
\newblock In \emph{Proceedings of the 2023 Conference on Empirical Methods in Natural Language Processing}, pages 2695--2709.

\bibitem[{Qin et~al.(2024{\natexlab{b}})Qin, Chen, Zhou, Chen, Li, Liao, Li, Che, and Yu}]{qin2024multilingual}
Libo Qin, Qiguang Chen, Yuhang Zhou, Zhi Chen, Yinghui Li, Lizi Liao, Min Li, Wanxiang Che, and Philip~S Yu. 2024{\natexlab{b}}.
\newblock \href {https://arxiv.org/abs/2404.04925} {Multilingual large language model: A survey of resources, taxonomy and frontiers}.
\newblock \emph{ArXiv preprint}, abs/2404.04925.

\bibitem[{Roy and Roth(2015)}]{roy-roth-2015-solving}
Subhro Roy and Dan Roth. 2015.
\newblock \href {https://doi.org/10.18653/v1/D15-1202} {Solving general arithmetic word problems}.
\newblock In \emph{Proceedings of the 2015 Conference on Empirical Methods in Natural Language Processing}, pages 1743--1752.

\bibitem[{Roy et~al.(2015)Roy, Vieira, and Roth}]{roy2015reasoning}
Subhro Roy, Tim Vieira, and Dan Roth. 2015.
\newblock \href {https://doi.org/10.1162/tacl_a_00118} {Reasoning about quantities in natural language}.
\newblock \emph{Transactions of the Association for Computational Linguistics}, Volume 3:1--13.

\bibitem[{Sel et~al.(2023)Sel, Tawaha, and Khattar}]{sel2023algorithm}
Bilgehan Sel, Ahmad Tawaha, and Vanshaj et~al. Khattar. 2023.
\newblock Algorithm of thoughts: Enhancing exploration of ideas in large language models.
\newblock In \emph{Forty-first International Conference on Machine Learning}.

\bibitem[{Team et~al.(2023)Team, Anil, Borgeaud, Wu, Alayrac, Yu, Soricut, Schalkwyk, Dai, Hauth et~al.}]{team2023gemini}
Gemini Team, Rohan Anil, Sebastian Borgeaud, Yonghui Wu, Jean-Baptiste Alayrac, Jiahui Yu, Radu Soricut, Johan Schalkwyk, Andrew~M Dai, Anja Hauth, et~al. 2023.
\newblock \href {https://arxiv.org/abs/2312.11805} {Gemini: a family of highly capable multimodal models}.
\newblock \emph{ArXiv preprint}, abs/2312.11805.

\bibitem[{Tong et~al.(2024)Tong, Li, and Wang}]{tong2024can}
Yongqi Tong, Dawei Li, and Sizhe et~al. Wang. 2024.
\newblock \href {https://arxiv.org/abs/2403.20046} {Can llms learn from previous mistakes? investigating llms' errors to boost for reasoning}.
\newblock \emph{ArXiv preprint}, abs/2403.20046.

\bibitem[{Touvron et~al.(2023)Touvron, Martin, Stone, Albert, Almahairi, Babaei, Bashlykov, Batra, Bhargava, Bhosale et~al.}]{touvron2023llama}
Hugo Touvron, Louis Martin, Kevin Stone, Peter Albert, Amjad Almahairi, Yasmine Babaei, Nikolay Bashlykov, Soumya Batra, Prajjwal Bhargava, Shruti Bhosale, et~al. 2023.
\newblock \href {https://arxiv.org/abs/2307.09288} {Llama 2: Open foundation and fine-tuned chat models}.
\newblock \emph{ArXiv preprint}, abs/2307.09288.

\bibitem[{Wei et~al.(2022)Wei, Wang, and Schuurmans}]{wei2022chain}
Jason Wei, Xuezhi Wang, and Dale et~al. Schuurmans. 2022.
\newblock Chain-of-thought prompting elicits reasoning in large language models.
\newblock \emph{Advances in neural information processing systems}, 35:24824--24837.

\bibitem[{Xu et~al.(2024{\natexlab{a}})Xu, Fei, Pan, Liu, Lee, and Hsu}]{xu2024faithful}
Jundong Xu, Hao Fei, Liangming Pan, Qian Liu, Mong-Li Lee, and Wynne Hsu. 2024{\natexlab{a}}.
\newblock \href {https://doi.org/10.18653/v1/2024.acl-long.720} {Faithful logical reasoning via symbolic chain-of-thought}.
\newblock In \emph{Proceedings of the 62nd Annual Meeting of the Association for Computational Linguistics (Volume 1: Long Papers)}, pages 13326--13365.

\bibitem[{Xu et~al.(2024{\natexlab{b}})Xu, Liu, Liu, and et~al.}]{xu2024activerag}
Zhipeng Xu, Zhenghao Liu, Yibin Liu, and et~al. 2024{\natexlab{b}}.
\newblock \href {https://arxiv.org/abs/2402.13547} {Active{RAG}: Revealing the treasures of knowledge via active learning}.
\newblock \emph{ArXiv preprint}, abs/2402.13547.

\bibitem[{Yao et~al.(2023)Yao, Yu, and Zhao}]{yao2024tree}
Shunyu Yao, Dian Yu, and Jeffrey et~al. Zhao. 2023.
\newblock \href {https://proceedings.neurips.cc/paper_files/paper/2023/file/271db9922b8d1f4dd7aaef84ed5ac703-Paper-Conference.pdf} {Tree of thoughts: Deliberate problem solving with large language models}.
\newblock In \emph{Advances in Neural Information Processing Systems}, volume~36, pages 11809--11822. Curran Associates, Inc.

\bibitem[{Zhang et~al.(2024{\natexlab{a}})Zhang, Madaan, Gao, Zhang, Mishra, Yang, Tandon, and Alon}]{zhang2024context}
Tianjun Zhang, Aman Madaan, Luyu Gao, Steven Zhang, Swaroop Mishra, Yiming Yang, Niket Tandon, and Uri Alon. 2024{\natexlab{a}}.
\newblock In-context principle learning from mistakes.
\newblock In \emph{ICML 2024 Workshop on In-Context Learning}.

\bibitem[{Zhang et~al.(2024{\natexlab{b}})Zhang, Chen, Li, Che, and Qin}]{zhang2024autocap}
Yongheng Zhang, Qiguang Chen, Min Li, Wanxiang Che, and Libo Qin. 2024{\natexlab{b}}.
\newblock \href {https://doi.org/10.18653/v1/2024.findings-acl.546} {Autocap: Towards automatic cross-lingual alignment planning for zero-shot chain-of-thought}.
\newblock pages 9191--9200.

\bibitem[{Zhang et~al.(2022)Zhang, Zhang, Li, and Smola}]{zhang2022automatic}
Zhuosheng Zhang, Aston Zhang, Mu~Li, and Alex Smola. 2022.
\newblock Automatic chain of thought prompting in large language models.
\newblock In \emph{The Eleventh International Conference on Learning Representations}.

\bibitem[{Zheng et~al.(2024)Zheng, Liu, Xie, Li, and Li}]{zheng2023progressive}
Chuanyang Zheng, Zhengying Liu, Enze Xie, Zhenguo Li, and Yu~Li. 2024.
\newblock \href {https://openreview.net/forum?id=UkFEs3ciz8} {Progressive-hint prompting improves reasoning in large language models}.
\newblock In \emph{AI for Math Workshop @ ICML 2024}.

\bibitem[{Zhou et~al.(2024)Zhou, Pujara, Ren, Chen, Cheng, Le, Chi, Zhou, Mishra, and Zheng}]{zhou2024self}
Pei Zhou, Jay Pujara, Xiang Ren, Xinyun Chen, Heng-Tze Cheng, Quoc~V Le, Ed~H Chi, Denny Zhou, Swaroop Mishra, and Huaixiu~Steven Zheng. 2024.
\newblock Self-discover: Large language models self-compose reasoning structures.
\newblock \emph{arXiv preprint arXiv:2402.03620}.

\end{thebibliography}

\appendix

\end{CJK}
\end{document}